\documentclass[conference, onecolumn]{IEEEtran}

\IEEEoverridecommandlockouts
\usepackage{amsmath,amssymb,amsfonts}
\usepackage{algorithmic}
\usepackage{graphicx}
\usepackage{textcomp}
\usepackage{xcolor}
\usepackage{natbib}
\usepackage{hyperref} 
\usepackage{array,booktabs}

\graphicspath{{./}} 
\def\BibTeX{{\rm B\kern-.05em{\sc i\kern-.025em b}\kern-.08em
    T\kern-.1667em\lower.7ex\hbox{E}\kern-.125emX}}

\begin{document}

\title{Survey of Imbalanced Data Methodologies\\
}

\author{\IEEEauthorblockN{Lian Yu}
\IEEEauthorblockA{\textit{Corporate Model Risk} \\
\textit{Wells Fargo, US}\\
lian.yu@wellsfargo.com} 
\and
\IEEEauthorblockN{Nengfeng Zhou}
\IEEEauthorblockA{\textit{Corporate Model Risk}\\
\textit{Wells Fargo, US}\\ 
nengfeng.zhou@wellsfargo.com}
}

\maketitle

  \def\mystrut(#1,#2){\vrule height #1pt depth #2pt width 0pt}    
  \def\arraystretch{1.3}

\begin{abstract}
Imbalanced data set is a problem often found and well-studied in financial industry. In this paper, we reviewed and compared some popular methodologies handling data imbalance. We then applied the under-sampling/over-sampling methodologies to several modeling algorithms on UCI and Keel data sets. The performance was analyzed for class-imbalance methods, modeling algorithms and grid search criteria comparison.

\end{abstract}


\section{Introduction}
A data set is called imbalanced if the classification classes are not approximately equal, where one class contains many more samples than the rest of the classes. In this scenario, classifiers can have good predictive accuracy on the majority class but poor performance on the minority class(es) due to the larger influence of majority class. Hence, the traditional modeling algorithms do not always perform well on the imbalanced data, and the class-imbalance methodologies are proposed to improve the prediction performance of the modeling algorithms.

In this paper, we review methodologies dealing with imbalanced data and the corresponding performance measures. We then evaluate the impact of class-imbalance methods on many traditional modeling algorithms with empirical experiments. The imbalanced data problem draws many attentions in literature and empirical works. Depending on the modeling stages applied on, the class-imbalance methodologies can be classified into data pre-processing methods and modeling algorithm specific methods. The data pre-processing methods are usually under/over-sampling methods that apply on the training data before modeling. The modeling algorithm specific methods are stand-alone algorithms that work on imbalanced training data directly. To evaluate the impact of class-imbalance methods on the model performance, 15 data sets from UCI/Keel databases are tested, where each data set contains at least 500 samples and the range of imbalance ratio is wide. We experiment four class-imbalance methods on eight modeling algorithms, and measure the performance with F-score and AUC. The results show that the choice of modeling algorithms has more impact on the performance, while the class-imbalance methods are more effective on simple linear algorithms like Logistic Regression and Linear SVC.

The paper is organized as follows. Section 2 describes the class-imbalance methods and the performance measure for imbalanced data. Section 3 compares the performance of imbalance methods and modeling algorithms through empirical experiments. 
 
\section{Class-imbalance methodologies and performance measures}
In this section, we present a literature review of techniques to handle imbalanced data sets, including sampling methods, algorithms of data ensemble and cost sensitive approaches. Furthermore, we present several performance measures for data imbalance, the way to select appropriate measures and their impacts on performance evaluation. 

\subsection{Sampling methods}

Resampling the original data set is a technique that applies at data level for balancing the majority and minority classes. The method works at data pre-processing step, either by over-sampling the minority class or by under-sampling the majority class, to construct a well-balanced training data set. After that, any modeling algorithm could be trained on such data set to alleviate the bias of the algorithm towards the majority class.

Random majority under-sampling method is the most basic statistical approach that randomly discards samples from the majority class. Since the samples from the majority classes are removed, this method can potentially ignore useful information from those removed samples. Therefore, several under-sampling approaches are proposed to selectively remove samples from the majority class so that the information could be largely retained in the training data set. Condensed Nearest Neighbor (CNN) rule \citep{hart68} is one of the first techniques that only removes majority samples far away from the decision neighbor. The CNN rule is repeated on the training data set until the set to be removed is stable. Edited Nearest Neighbor (ENN) rule \citep{wilson72} uses nearest neighbor for removing samples that do not agree with the majority of its k nearest neighbors. This algorithm edits out the samples that are identified as noise or borderline, and leaves smoother decision boundaries. Tomek Links \citep{tomek76} is another approach such that only majority samples identified as Tomek Links are removed. By checking Tomek Links between nearest neighbor pairs, majority samples are removed until all minimally distanced nearest neighbor pairs are in the same class. Near Miss \citep{zhang03} is a family of under-sampling techniques that remove majority samples based on their average distances from the minority class. Depending on the nearest neighbor algorithms used to measure the distance, three Near Miss methods are proposed. Near Miss-1 selects majority samples with smallest average distance to three closest samples from the minority class. Near Miss-2 selects majority samples with smallest average distance to three farthest samples from the minority class. Near Miss-3 selects the k closest majority samples for each sample of the minority class. In practice, the under-sampling methods could also be used in combination to further improve data imbalance. One-Sided Selection \citep{kubat97} is such a hybrid algorithm, where the CNN rule reduces the majority samples by keeping the samples in a sub-set with the one-nearest neighbor rule (1-NN), then the borderline or noisy samples that detected by Tomek Links are further removed. Neighborhood cleaning rule \citep{laurikkala01} works similarly as the one-sided selection by applying CNN rule then Wilson's ENN rule to identify noisy samples.

Over-sampling is another set of efficient sampling techniques to handle class imbalance, which artificially increases samples from the minority class while keeping all the majority samples. In that case, the information of the majority class is fully retained. Random minority over-sampling is such an approach that randomly duplicates minority samples and adds them to the training data set. However, even though the imbalance ratio is improved with this technique, duplicating the minority samples leads the modeling algorithm trained more on specific regions, or over-fitting. Therefore, several synthetic over-sampling approaches are proposed to increase the variety of minority class and reduce learning bias. Synthetic Minority Over-sampling Technique (SMOTE) \citep{chawla02} generates synthetic samples for minority class based on their nearest neighbors to shift the learning bias toward the minority class. The algorithm chooses nearest neighbor by Euclidean distance between data points and generates the synthetic samples by taking a linear segment between the sample under consideration and its nearest neighbor. Based on the regular SMOTE algorithm, extensions with different distance measures or selection of samples in consideration are proposed. For instance, in borderline SMOTE \citep{han05}, only minority samples near the borderline are over-sampled. In safe-level SMOTE \citep{bunkhumpornpat09}, minority samples along the same line with different weight degree, called safe level, are over-sampled. In density-based SMOTE \citep{bunkhumpornpat11}, only monitory samples from density-based notion of clusters are over-sampled. In addition to SMOTE algorithms, Adaptive Synthetic \citep{he08} sampling is an over-sampling approach that uses a weighted distribution for different minority samples according to their level of difficulty in learning. In AdaSyn algorithm, more synthetic data is generated for minority samples that are harder to learn compared to those minority samples that are easier to learn.

Both over-sampling and under-sampling techniques can be effective when used in isolation, while combination of both techniques could be more effective for imbalanced data sets. SMOTE + Tomek Links and SMOTE + ENN \citep{batista04} are proposed to address the concern of over-fitting introduced by artificial minority samples. First, the SMOTE algorithm is applied to the original data set to over-sample the minority samples. Then, majority samples in Tomek Links are identified and removed, or the misclassified majority samples by its nearest neighbors is removed by the ENN rule. Both hybrid algorithms have been showed good experiment results when compared with the over-sampling methods alone.

\subsection{Ensemble classifiers and cost sensitive approaches}

The over- and under-sampling approaches are not specific to any modeling algorithms, which could be considered as data pre-processing steps. Data imbalance problem could also be handled by ensemble classifiers and cost sensitive approaches, which are algorithm specific and could be considered as algorithm enhancement. AdaBoost \citep{freund96}, which ensembles of weak learners on various distributions, is one of the early successful boosting algorithms that work for classification learning problem. Since then, variants of hybrid sampling/boosting algorithms are proposed for imbalanced data sets. SMOTEBoost \citep{chawla03} iteratively trains the AdaBoost learners on balanced subsets with synthetic minority samples generated by SMOTE, and combines the outputs of those learners. On the contrary, Random under-sampling boost (RUSB) \citep{seiffert10} iteratively trains the AdaBoost learners on the randomly under-sampled data set. The RUSB greatly overcomes the main drawback of random under-sampling, which is the loss of information, by combining it with boosting. Balance Cascade approach \citep{liu09} trains the AdaBoost learners sequentially, where correctly classified majority samples are removed from training data sets in each iteration. Besides the boosting algorithms, Balanced Random Forest \citep{chen04} trains individual trees on more balanced subsets with bootstrap samples of the minority class and randomly selected majority samples, and takes a majority vote for prediction.

The cost sensitive approaches, which penalize more on misclassification of the minority class, have also been reported to be effective to handle class imbalance problem. AdaCost \citep{sun05} is an AdaBoost learner with more weights on the misclassified minority samples and updates the sample weights in the training process. Easy ensemble (Easy) \citep{liu09} trains AdaBoost learners on balanced subsets with majority samples randomly under-sampled, and intentionally increases the weights of minority samples. Weighted Random Forest \citep{chen04} assigns higher misclassification cost on the minority class, and uses the class weight to find splits and classify the terminal leaves when growing the tree.

\subsection{Performance measures of imbalanced data}

We have discussed the techniques that deal with data imbalance in the training data, and we notes that the evaluation of model performance is based on the testing data. Although the training data are sampled through the imbalance methods, the testing data are not sampled and its class distribution are not the same as the sampled training data. In this section, we review the common performance measures and their appropriateness to imbalanced data. 

For binary classification problem, the confusion matrix defines the base for performance measures. Most of the performance metrics are derived from the confusion matrix, for example, accuracy, misclassification rate, precision and recall. Accuracy is a measure of the overall efficiency of a model and is defined as:
\begin{equation*}
\text{Accuracy} = \frac{\text{Correct Predictions}}{\text{Total Predictions}} =  \frac{\text{True Positive + False Negative}}{\text{True Positive + False Positive + True Negative + False Negative}}
\end{equation*}
However, the accuracy may not be appropriate when the data is imbalanced. In that case, more weights are placed on the majority class than on the minority class, and the model may not perform well on the minority class even with a high accuracy.

To accomodate the cost of the minority class, Receiver Operating Characteristic (ROC) \citep{swets88} curve is proposed as a measure over a range of tradeoffs between the True Positive Rate and False Positive Rate. Area Under the Curve (AUC) is a commonly used performance metric for summarizing the ROC curve in a single score. Moreover, AUC is not biased toward model's performance on the majority or minority class, which makes this measure more appropriate when dealing with imbalanced data.

From the confusion matrix, we can also derive precision and recall \citep{buckland94} performance metrics, which are defined as:
\begin{equation*}
\text{Precision} = \frac{\text{True Positive}}{\text{True Positive + False Positive}}
\end{equation*}
and
\begin{equation*}
\text{Recall} = \frac{\text{True Positive}}{\text{True Positive + False Negative}}
\end{equation*}
For imbalanced data, the main goal is to improve the True Positive for the minority class, however, the number of False Positives can also increase in that case. To balance the recall and precision, i.e. improving the recall, while keeping precision low, the F-score \citep{buckland94} is proposed as a harmonic mean of the precision and recall:
\begin{equation*}
F\text{-score} = 2\cdot\frac{\text{Precision}\cdot\text{Recall}}{\text{Precision + Recall}} 
\end{equation*}
Since the F-score weights precision and recall equally and balances both concerns, it is less likely to be biased to the majority or minority class.

\section{Comparison of Class-imbalance Methods}
\subsection{Experimental Settings}
We test the imbalance methods on 15 data sets. The data sets are from UCI \citep{uci20} and Keel \citep{keel11} databases and we further randomly down-sample the minority samples to achieve a higher imbalance ratio for some data sets. The statistics of these data sets with imbalance ratio ranging from 9 to 130 are summarized in Table  \ref{tab:addlabel0}. 

\begin{table}[htbp]
  \centering
  \caption{Summary Statistics of Experimental Data Sets}
    \begin{tabular}{l|r|r|r|r|r}
    \hline
    \textbf{Data} & \textbf{Size} & \textbf{Minority/Majority} & \textbf{Ratio} & \textbf{Data Source} & \textbf{Original Data Size}\\
    \hline
    
    \textbf{Adult} & 25,503  & 784/24,719 &  31.5 & UCI	& 7,841/24,719 \\
    \textbf{Abalone2} &  3,864  &  78/3,786  &  48.5 & UCI & 391/3,786 \\
    \textbf{Car eval} & 1,421  & 77/1,344  &  17.5 & UCI & 384/1,344 \\
    \textbf{Wifi\_localization} & 1,550 &  50/1,500  & 30 & UCI & 500/1,500 \\
    \textbf{Satimage} & 5,073 & 151/4,922  & 32.6 & UCI & 1,508/4,922 \\
    \textbf{Wine} & 5,058 & 160/4,898 & 30.6 & UCI & 1,599/4,898 \\
    \textbf{Letter-recognition2} & 20,000 & 789/19,211 & 24.3 & UCI & 789/19,211 \\
    \hline
    \textbf{Yeast 3} &  1,354 &  33/1,321 &  40.0 & Keel & 163/1,321 \\
    \textbf{Pima} & 554 & 54/500  &  9.3 & Keel & 268/500 \\
    \textbf{Abalone\_19} & 4,174 & 32/4,142 & 129.4 & Keel & 32/4,142 \\
    \textbf{Page-blocks0} &  4,969 & 56/4,913  & 87.7 & Keel & 559/4,913 \\
    \textbf{Yeast-0-2-5-6\_vs\_3-7-8-9} & 1,004 & 99/905 & 9.1 & Keel & 99/905 \\
    \textbf{Yeast-0-2-5-7-9\_vs\_3-6-8} & 1,004 & 99/905  & 9.1 & Keel & 99/905 \\
    \textbf{Yeast6} & 1,484 & 35/1,449 &41.4 & Keel & 35/1,449 \\
    \textbf{Yeast1} & 1,098 & 43/1,055 & 24.5 & Keel & 429/1,055 \\
    \hline
    \end{tabular}%
  \label{tab:addlabel0}%
\end{table}%

For each data set, we randomly split the sample into training and testing data sets on a 70:30 split before model development. The target is a binary variable labeled with 1 for minority samples and 0 for majority samples. The model performance is measured by F-score and AUC on testing data sets. The model score cutoffs are based on scores of original testing data sets. The scores are rank-ordered from highest to lowest, and the cutoff is determined such that the percentage of predicted 1's is the same as percentage of actual 1's. The whole development/testing process is repeated for 50 times, and the performance values are averaged over the 50 iterations for each data set. The final performance value at algorithm level is the averaged F-score or AUC over the 15 experimental data sets. 

We compare the performance of eight algorithms in this experiment, which include Logistic Regression (LR), Linear Support Vector Classifier (Linear SVC), Nearest Neighbor (NN), Classification and Regression Trees (CART), Random Forests (RF), Gradient Boosting Tree (XGBoost), Under-sampling with AdaBoost (RUSB) \citep{seiffert10} and Easy Ensemble (Easy) \citep{liu09}. For each modeling algorithm, the models are built on the original training data sets as the baseline and on the training data sets after applying imbalance methods for comparison. We test four class-imbalance methods: SMOTE, Adaptive Synthetic Sampling (AdaSyn), SMOTE-ENN and SMOTE-Tomek. The hyper-parameters for these modeling algorithms are tuned by grid search so that F-score, AUC and accuracy metrics on training data sets are optimized, respectively. Finally, the model performance across modeling algorithms and imbalance methods is evaluated by the average F-score or AUC on original testing data sets.

\subsection{Analysis of Performance}

The performance of imbalance methods for the eight modeling algorithms has been summarized in tables below. Table \ref{tab:addlabel1}, \ref{tab:addlabel2} and \ref{tab:addlabel3} have presented the average F-score and AUC for the models that are tuned by grid search using AUC, F-score, and accuracy, respectively. We evaluate the performance in terms of class-imbalance methods, modeling algorithms and grid search criteria.

\begin{table}[htbp]
  \centering
  \caption{Average Performance with Grid Search by AUC}
    \begin{tabular}{p{0.12\textwidth}|p{0.07\textwidth}|p{0.08\textwidth}|p{0.08\textwidth}|p{0.08\textwidth}|p{0.07\textwidth}|p{0.07\textwidth}|p{0.07\textwidth}|p{0.07\textwidth}|p{0.05\textwidth}}
    \hline 
    {} & \multicolumn{2}{c|}{\textbf{Linear Algorithms}} & \multicolumn{2}{c|}{\textbf{Simple Non-linear Algorithms}} & \multicolumn{2}{c|}{\textbf{Ensemble Algorithms}} & \multicolumn{2}{c|}{\textbf{Imbalance Algorithms}} &\\ 
    \hline
    \textbf{F-score} & \textbf{LR} & \textbf{Linear SVC} & \textbf{NN} & \textbf{CART} & \textbf{RF} & \textbf{XGBoost} & \textbf{RUSB} & \textbf{Easy} & \textbf{Average}\\
    \hline
    \textbf{None}  & \textbf{0.6781} & 0.6625 & \textbf{0.6909} & \textbf{0.6469} & \textbf{0.7090} & 0.7036 & 0.6847 & \textbf{0.6901} & \textbf{0.6832} \\ 		
    \textbf{SMOTE} & 0.6750 & 0.6646 & 0.6484 & 0.6147 & 0.6919 & \textbf{0.7099} & 0.6886 & 0.6650 & 0.6698 \\   								
    \textbf{AdaSyn} & 0.6681 & 0.6591 & 0.6083 & 0.5944 & 0.6787 & 0.7016 & 0.6829 & 0.6531 & 0.6558 \\ 								
    \textbf{SMOTE-ENN} & 0.6765 & \textbf{0.6666} & 0.5057 & 0.5530 & 0.6887 & 0.7036 & \textbf{0.6905} & 0.6721 & 0.6446\\ 								
    \textbf{SMOTE-Tomek} & 0.6746 & 0.6651 & 0.6482 & 0.6097 & 0.6936 & 0.7093 & 0.6905 & 0.6668 & 0.6697\\ 
    \hline
    \textbf{AUC} & \textbf{LR} & \textbf{Linear SVC} & \textbf{NN} & \textbf{CART} & \textbf{RF} & \textbf{XGBoost} & \textbf{RUSB} & \textbf{Easy} & \textbf{Average} \\
    \hline
    \textbf{None}  & 0.8508 & 0.8333 & 0.8079 & \textbf{0.8163} & 0.8941 & 0.8935 & \textbf{0.8934} & \textbf{0.8913} & 0.8601\\								
    \textbf{SMOTE} & 0.8777 & 0.8597 & \textbf{0.8378} & 0.8037 & 0.8957 & 0.8968 & 0.8719 & 0.8741 & \textbf{0.8647}\\ 								
    \textbf{AdaSyn} & 0.8761 & 0.8583 & 0.8350 & 0.8124 & 0.8911 & 0.8952 & 0.8676 & 0.8688 & 0.8631\\ 								
    \textbf{SMOTE-ENN} & 0.8775 & \textbf{0.8624} & 0.8081 & 0.8087 & \textbf{0.8966} & \textbf{0.8981} & 0.8790 & 0.8817 & 0.8640\\ 								
    \textbf{SMOTE-Tomek} & \textbf{0.8783} & 0.8599 & 0.8369 & 0.8023 & 0.8961 & 0.8966 & 0.8717 & 0.8751 & 0.8646 \\ 
    \hline
    \end{tabular}%
  \label{tab:addlabel1}%
\end{table}%

\begin{table}[htbp]
  \centering
  \caption{Average Performance with Grid Search by F-score}
    \begin{tabular}{p{0.12\textwidth}|p{0.07\textwidth}|p{0.08\textwidth}|p{0.08\textwidth}|p{0.08\textwidth}|p{0.07\textwidth}|p{0.07\textwidth}|p{0.07\textwidth}|p{0.07\textwidth}|p{0.05\textwidth}}
    \hline
    {} & \multicolumn{2}{c|}{\textbf{Linear Algorithms}} & \multicolumn{2}{c|}{\textbf{Simple Non-linear Algorithms}} & \multicolumn{2}{c|}{\textbf{Ensemble Algorithms}} & \multicolumn{2}{c|}{\textbf{Imbalance Algorithms}} &\\
    \hline
    \textbf{F-score} & \textbf{LR} & \textbf{Linear SVC} & \textbf{NN} & \textbf{CART}  & \textbf{RF} & \textbf{XGBoost} & \textbf{RUSB} & \textbf{Easy} &\textbf{Average} \\
    \hline
    \textbf{None}  & 0.6725 & 0.6417 & \textbf{0.6891} & \textbf{0.6665} & \textbf{0.7077} & \textbf{0.7178} & 0.6854 & \textbf{0.6881} & \textbf{0.6836} \\ 									
    \textbf{SMOTE} & 0.6766 & 0.6928 & 0.6115 & 0.6063 & 0.6642 & 0.7126 & 0.6889 & 0.6647 & 0.6647 \\   								
    \textbf{AdaSyn} & 0.6703 & 0.6597 & 0.5729 & 0.5892 & 0.6768 & 0.7095 & 0.6813 & 0.6529 & 0.6516 \\ 								
    \textbf{SMOTE-ENN} & 0.6739 & 0.6650 & 0.5054 & 0.5387 & 0.6907 & 0.7027 & \textbf{0.6913} & 0.6739 & 0.6427\\ 								
    \textbf{SMOTE-Tomek} & \textbf{0.6773} & \textbf{0.6653} & 0.6070 & 0.6031 & 0.6940 & 0.7124 & 0.6887 & 0.6661 & 0.6642\\ 
    \hline
    \textbf{AUC} & \textbf{LR} & \textbf{Linear SVC} & \textbf{NN} & \textbf{CART} & \textbf{RF} & \textbf{XGBoost} & \textbf{RUSB} & \textbf{Easy} & \textbf{Average}\\
    \hline
    \textbf{None}  & 0.8387 & 0.8084 & 0.7784 & 0.7685 & 0.8927 & 0.8778 & \textbf{0.8930} & \textbf{0.8901} & 0.8434\\								
    \textbf{SMOTE} & \textbf{0.8777} & 0.8599 & 0.8227 & 0.7900 & 0.8954 & 0.8839 & 0.8714 & 0.8738 & 0.8594\\ 								
    \textbf{AdaSyn} & 0.8763 & 0.8577 & \textbf{0.8248} & 0.7987 & 0.8917 & 0.8835 & 0.8676 & 0.8693 & 0.8587\\ 								
    \textbf{SMOTE-ENN} & 0.8727 & 0.8571 & 0.8037 & \textbf{0.7987} & \textbf{0.8963} & \textbf{0.8901} & 0.8793 & 0.8823 & \textbf{0.8600} \\ 								
    \textbf{SMOTE-Tomek} & 0.8775 & \textbf{0.8613} & 0.8224 & 0.7884 & 0.8957 & 0.8844 & 0.8706 & 0.8745 & 0.8594 \\ 				
    \hline
    \end{tabular}%
  \label{tab:addlabel2}%
\end{table}%

\begin{table}[htbp]
  \centering
 \caption{Average Performance with Grid Search by Accuracy}
    \begin{tabular}{p{0.14\textwidth}|p{0.08\textwidth}|p{0.08\textwidth}|p{0.08\textwidth}|p{0.08\textwidth}|p{0.08\textwidth}|p{0.08\textwidth}|p{0.05\textwidth}}
    \hline
    {} & \multicolumn{2}{c|}{\textbf{Linear Algorithms}} & \multicolumn{2}{c|}{\textbf{Simple Non-linear Algorithms}} & \multicolumn{2}{c|}{\textbf{Ensemble Algorithms}} &\\
    \hline
    \textbf{F-score} & \textbf{LR} & \textbf{Linear SVC} & \textbf{NN} & \textbf{CART} & \textbf{RF} & \textbf{XGBoost} & \textbf{Average}\\
    \hline    
    \textbf{None}  & 0.6705 & 0.6421 & \textbf{0.6877} & \textbf{0.6596} & \textbf{0.7073} & 0.7030 & \textbf{0.6784}\\
    \textbf{SMOTE} & \textbf{0.6763} & 0.6638 & 0.6120 & 0.6043 & 0.6932 & \textbf{0.7098} & 0.6599\\
    \textbf{AdaSyn}& 0.6702 & 0.6591 & 0.5729 & 0.5910 & 0.6798 & 0.7021 & 0.6458\\
    \textbf{SMOTE-ENN} & 0.6749 & 0.6639 & 0.5053 & 0.5389 & 0.6902 & 0.7023 & 0.6293 \\
    \textbf{SMOTE-Tomek} & 0.6759 & \textbf{0.6651} & 0.6054 & 0.6035 & 0.6929 & 0.7095 & 0.6587\\
    \hline
    \textbf{AUC} & \textbf{LR} & \textbf{Linear SVC} & \textbf{NN} & \textbf{CART} & \textbf{RF} & \textbf{XGBoost} & \textbf{Average}\\
    \hline
    \textbf{None} & 0.8379 & 0.8108 & 0.7849 & 0.7963 & 0.8933 & 0.8911 & 0.8357 \\
    \textbf{SMOTE} & 0.8772 & 0.8591 & 0.8231 & 0.7904 & 0.8956 & 0.8965 & 0.8570\\
     \textbf{AdaSyn} & 0.8763 & 0.8570 & \textbf{0.8251} & \textbf{0.7992} & 0.8916 & 0.8955 & \textbf{0.8574} \\
    \textbf{SMOTE-ENN} & 0.8733 & 0.8573 & 0.8023 & 0.7991 & \textbf{0.8963} & \textbf{0.8984} & 0.8544\\
    \textbf{SMOTE-Tomek} & \textbf{0.8774} & \textbf{0.8599} & 0.8223 & 0.7899 & 0.8957 & 0.8966 & 0.8570\\
    \hline
    \end{tabular}%
  \label{tab:addlabel3}%
\end{table}%

First, we compare the impact of imbalance methods across modeling algorithm categories. Since the performance is similar over grid search criteria and is evaluated by F-score and AUC, we focus on results in Table  \ref{tab:addlabel1}.

1) For the linear algorithms, Logistic Regression and Linear SVC, the model performance measured by AUC improves after applying the imbalance methods to the training data sets. With any of the class-imbalance methods, the AUC improves significantly from 85.08\% to 87\% for Logistic Regression, and from 83.33\% to 86\% for Linear SVC. The F-score is not very sensitive to the imbalance methods, where the values are comparable with the original baseline for SMOTE methods, and drop slightly for Adaptive Synthetic method. Overall, both linear algorithms perform better with imbalance methods, and it is recommended to pre-process the data sets before training the model.

2) For simple non-linear algorithms, Nearest Neighbor and CART, the performance results are mixing and we do not observe consistent improvement. For Nearest Neighbor, the AUC improves from 80.79\% to 83\% for most of the imbalance methods except only 0.02\% improvement for SMOTE-ENN. For CART, the AUC is the highest when the model is trained on the original data set, and decreases slightly after applying imbalance methods. When measured by F-score, the performance drops significantly from original 69.09\% to 50-65\% for Nearest Neighbor, and from 64.69\% to 55-61\% for CART. Specifically, with SMOTE-ENN, the F-score drops to 50.57\% for Nearest Neighbor and 55.30\% for CART. Since the Nearest Neighbor algorithm is mostly used to classify the new data point based on how its neighbors are classified, the decision boundary is sensitive to the sample composition. CART has the well-known weakness of non-robustness and a small change in the training data can result in a large change in the tree and consequently the final predictions. Therefore, both algorithms are more likely to be over-fitted when the minority class is over-sampled, and the performance on the more imbalanced testing data sets drops. We would not recommend applying imbalance methods for these simple non-linear algorithms.

3) For ensemble algorithms, Random Forest and XGBoost, the improvement of model performance is also not obvious after applying imbalance methods. For both algorithms, the AUCs with or without imbalance methods are around 89\% with few variation. The F-score for Random Forest decreases from the baseline 70.90\% to 67-69\% when pre-processed by imbalance methods, but for XGBoost, the values are almost the same. Since the ensemble algorithms already perform well on the imbalanced data sets, the impact of imbalance methods are not significant.

4) The RUSB and Easy Ensemble are ensemble algorithms that could handle data imbalance, where the AdaBoost model is trained iteratively on balanced samples with over-sampling or under-sampling. Since these two algorithms already re-balance the data in the training process, the additional imbalance methods in the data pre-processing step will not contribute further improvement to algorithm performance. For AUC, both algorithms perform best when trained on the original data sets. For F-score, we observe $<$ 1\% improvement for RUSB and slightly worse value for Easy Ensemble. 

The second set of comparison is conducted between modeling algorithms on imbalanced data sets. Still, the results are based on the grid search with AUC. 

1) For linear algorithms and simple non-linear algorithms, the AUC is greatly lower than more complex algorithms like ensemble algorithms and imbalance algorithms. The simpler algorithms tend to have better model interpretability than the more complex algorithms with scarificing on performance. Specifically, the more complex algorithms combine the decisions from multiple model iterations and improve the overall model performance even with data imbalance. The simple non-linear algorithms perform worst among all algorithm categories in AUC, and we would not recommend choosing these algorithms for imbalanced data sets.

2) When measured by F-score, the ensemble algorithms (Random Forest and XGBoost) outperform other algorithm categories. It is showed that with similar AUCs, the ensemble algorithms achieve slightly better F-score than imbalance algorithms (RUSB and Easy Ensemble). Therefore, even without imbalance methods, the ensemble algorithms still work well for imbalanced data sets.

Lastly, we compare the performance between grid search criteria. All eight modeling algorithms are tuned by grid search with AUC and F-score as criterion, and RUSB and Easy Ensemble could not be tuned by accuracy. 

1) The results are very close for different grid search criteria. Since the performance is measured by AUC and F-score, when the hyper-parameters are tuned by AUC, the algorithm performs better than other grid search criteria for the corresponding performance measurement. However, F-score is not consistently better when tuned by F-score.

2) The simple non-linear algorithms (Nearest Neighbor and CART) are more sensitive to grid search criterion. The more complex ensemble and imbalance algorithms are less sensitive and achieve high AUC and F-score for all three grid search criteria.

In summary, the linear algorithms perform better after applying imbalance methods in terms of AUC, without hurting F-score. For the more complex ensemble and imbalance algorithms, since these algorithms are not that sensitive to class imbalance, and already achieves high AUC score and F-Score, the improvement after applying imbalance methods is not so obvious. For simple non-linear algorithms (Nearest Neighbor and CART), applying class-imbalance methods on training data sets over fits the models and the results on testing data do not have consistent performance improvement measured by both AUC and F-score, compared to the baseline values. Our results also show that the four different imbalance methods tested have similar performance. There is no strong evidence to prefer one imbalance method over the other based on the comparison.

\section{Conclusion}

This paper provided an overview of different methodologies used to deal with imbalanced data. In the past 20 years, there are so many methodologies developed to deal with this problem since the publication of SMOTE \citep{chawla02}. It becomes difficult to choose the right imbalance methods in a real problem. Our empirical study compared the performance of some typical  imbalance methods on large number of real datasets. We also studied the interactions between class-imbalance methods and modeling algorithms. We found that imbalance methods do not always have consistent improvements. Their improvements are depending on which modeling algorithm is used. The imbalance methods are most suitable for simple linear algorithms. We also observed that different imbalance methods do not have significant different behaviors. Using more complicated ensemble modeling algorithms will achieve best performance on imbalanced data, even without applying imbalance methods. If AUC or F-score is the main objective of the model, we recommend to use more complicated ensemble modeling algorithms without imbalance methods. If model interpretation is also a key objective, we recommend to use linear algorithms combined with any of the imbalance methods studied in the paper.

\section{Declaration of Interest}

The authors report no conflicts of interest. The authors alone are responsible for the content and writing of the paper.

\section{Acknowledgment}
We thank Singhal Harsh for useful discussion. We thank corporate risk - model risk at Wells Fargo for support.
\bibliography{./main}
\bibliographystyle{apalike}

\end{document}